# Moving Vehicle Detection Using AdaBoost and Haar-Like Feature in Surveillance Videos

**Mohammad Mahdi Moghimi[1], Maryam Nayeri[2*], Majid Pourahmadi[3] and Mohammad Kazem Moghimi[4]**

[1,2,3]Department of Electrical Engineering, Yazd Branch, Islamic Azad University, Yazd, Iran.
Email: ms.moghimi@iauyazd.ac.ir; nayeri@iauyazd.ac.ir; pourahmadi@iauyazd.ac.ir,
[4]Department of Electrical Engineering, Shahed University, Tehran, Iran.
Email: kazem.moghimi@shahed.ac.ir

**ABSTRACT**

*Vehicle detection is a technology which its aim is to locate and show the vehicle size in digital images. In this technology, vehicles are detected in presence of other things like trees and buildings. It has an important role in many computer vision applications such as vehicle tracking, analyzing the traffic scene and efficient traffic management. In this paper, vehicles detected based on the boosting technique by Viola Jones. Our proposed system is tested in some real scenes of surveillance videos with different light conditions. The experimental results show that the accuracy, completeness, and quality of the proposed vehicle detection method are better than the previous techniques (about 94%, 92%, and 87%, respectively). Thus, our proposed approach is robust and efficient to detect vehicles in surveillance videos and their applications.*

## 1. INTRODUCTION

One of the usages of computer vision is to analyze the traffic scenes, such as counting vehicle and registering the driving violation like speed and illegal overtaking. In this case, we can point to some problems such as the variety of vehicles, the variety in size and color of vehicles and vehicle transforming in respect to position and vehicle distance in ratio of camera. Different algorithms are used to detect vehicles. According to the Hypothesis Generation (HG) features where the locations of candidate vehicles in an image are hypothesized. Based on Zehang Sun *et al* approach, can be classified the vehicle detection into one of the following three categories:

- Knowledge-based
- Stereo vision-based
- Motion-based

Knowledge-based methods are based on a prior knowledge of vehicle locations in an image by using features of vehicles such as:

- Symmetry; vehicles observed from rear or frontal view are in general symmetrically horizontal and vertical directions. This observation was used as a feature for vehicle detection.

- Color; although a lot of systems use color features to vehicle detection, but this feature is a very useful for detection of obstacle, tracking of the lane and road, or segmentation of vehicles and moving objects from background images.
- Shadow; assumed that image intensity, it was found that the region of the vehicle underneath is darker than any other regions on an asphalt paved road. It should be noted that the assumption might not always true. For example, rainy weather and bad illumination conditions will make the color of road pixels dark.
- Corners; assumed that in general have a rectangular shape.
- Horizontal and vertical edges; in different kind of vehicle views, especially rear views, they contain many horizontal and vertical lines, such as rear-window and bumper.
- Texture; the vehicles structure in an image causes local intensity changes. Due to similarities among all kind of vehicles, the intensity changes follow a certain pattern which is referred to the texture. This feature can be used as a narrow-down of the searched area for vehicle detection.
- Vehicle lights; we should assumed that the most of the features which used for vehicle detection, are not useful for night time. Vehicle lights represent a salient feature at night.

In stereo vision-based methods, there are two types of methods for vehicle detection. One of them uses disparity map and the other one uses an anti-perspective transformation (i.e., Inverse Perspective Mapping). The most common vehicle detection algorithm in urban highway is motion estimation algorithm based on Gaussian Mixture Model (GMM), optical flow and moving average technique. Large error is one of the disadvantages of this recent method. Because there are a lot of moving objects like motorcycles, commercial televisions, movement of tree by wind and movement of street shadow with time, so it is better to use the unchanged feature of the vehicles like vertical and horizontal edge, front and back lights, back angle of vehicle and the other things. Moghimi *et al.* has detected moving vehicles by combination of median filter and Principal Component Analysis (PCA) in surveillance videos. The moving objects are detected by moving average algorithm. And so, moving vehicles are recognized from other moving objects by PCA.

According to the Hypothesis Verification (HV) features which experimental results have been performed to verify the presence of a vehicle in an image and correctness of a hypothesis, HV approaches can be classified into two categories:
- Template-based methods
- Appearance-based methods

Template-based methods use predefined patterns of the vehicles classes and operate based on correlation between the image and the template, however, appearance-based methods learn the characteristics of the vehicles classes from a database of training images which have recorded the variability in vehicle appearance. Usually, the variability of the non-vehicle classes is also modeled to improve the performance of vehicle detection.

Therefore, this paper is introducing a method in order to vehicle detection by the feature of Haar-like filter. The paper is organized as follows; in section 2, our proposed method for moving car detection based on Viola Jones algorithm and Haar-Like feature is presented. Section 3 presents the preparing

XML files for vehicle detection. In section 4, experimental results are presented and analyzed. And finally, some discussions and conclusions are given in Section 5.

## 2. PROPOSED METHOD: MOVING CAR DETECTION BASED ON VIOLA JONES ALGORITHM AND HAAR-LIKE FEATURE

Object detection by Viola Jones algorithm is a real-time process. In this method, four key elements are AdaBoost, Haar-Like feature, cascaded classifier and integral image.

### 2.1. Haar-Like Feature

Haar-Like is a rectangular simple feature that is used as an input feature for cascaded classifier. In Fig. 1, there are some filters based on Haar-Like feature. By applying every one of these filters into one special area of the image, the pixel sums under white areas are subtracted from the pixel sums under the black areas. That is the weight of white and black area can be considered as "1" and "-1", respectively.

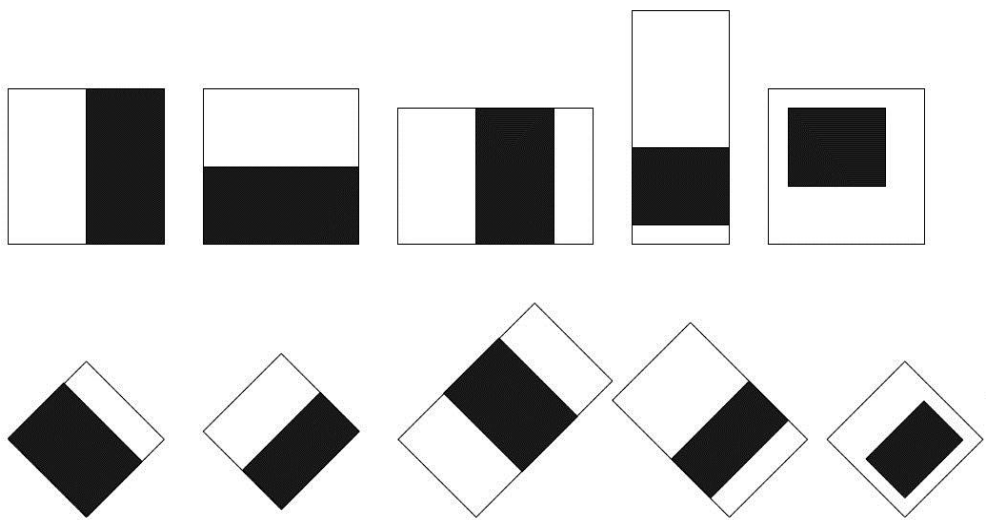

**Figure 1.** Different kinds of filters based on Haar-Like feature.

For example, for using shadow under the vehicle and tire position feature, spatial filters are used to extract the best feature. In Fig. 2, an example of these filters is shown.

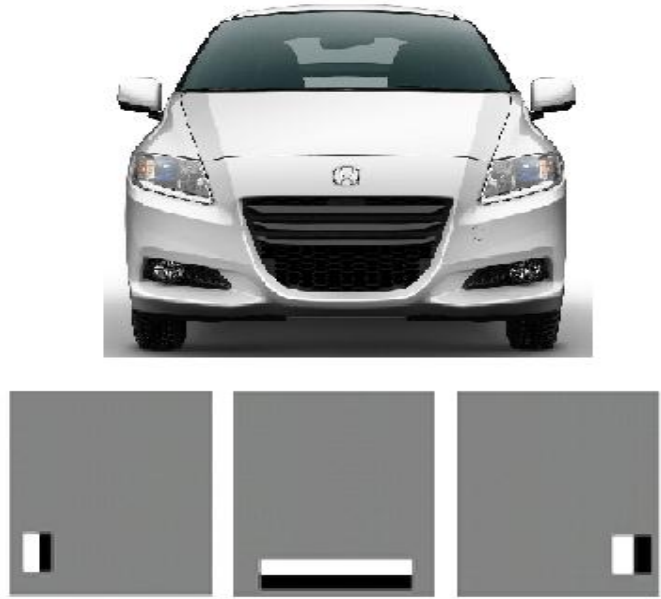

**Figure 2.** The used vehicle's features by Haar-Like filter.

### 2.2. AdaBoost Algorithm

AdaBoost algorithm (a machine learning meta-algorithm) in choosing features and improving the performance is repeatedly used. AdaBoost in order to construct a strong classifier combined many weak classifiers. Viola Jones algorithm uses the AdaBoost in the way that mixes a series of AdaBoost classifiers as a filter chain. Each filter is a separate AdaBoost classifier which consists of a few weak classifiers. If each of these filters in the acceptance region of the image shows the vehicle fails, this area is immediately classified as a non-vehicle. When a filter accepts an area of image as a vehicle, the area enters the next filter in the chain. If this area of the image passes all the chain filters successfully, it is classified as vehicle. In this algorithm, each cycle of boosting a feature among all other potential features is selected and in the end, the final classification will be a linear combines of the initial weak classification.

### 2.3. Integral Image

Integral image is a quick method for calculating the Haar-Like feature. Viola & Jones has used this technique and they recognized which Haar-Like feature among the other Haar-Likes is in each image. Integral image is sum of all pixel values in above and left of the position (x,y). Black area of the Fig. 3 shows the integral image. With integral image Haar-Like feature can be calculated quickly by simple addition and subtraction.

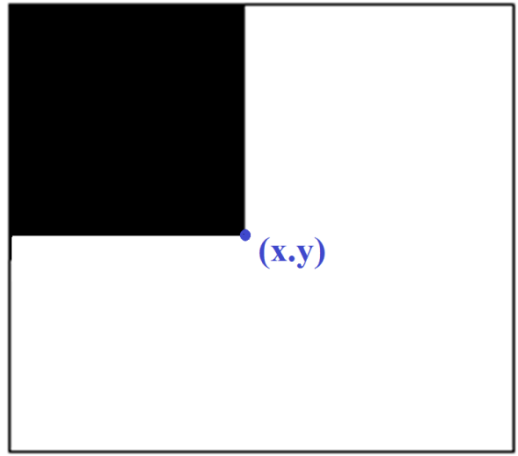

**Figure 3.** Integral image of the point of (x,y).

## 2.4. Cascaded Classifier

Cascaded classifier is used for fast rejection of error windows and improving the processing speed. In every node of trees there is a non-vehicle branching, it means that the image will not be vehicle. By this technique the false negative rate is at least.

For using Viola Jones algorithm in vehicle detection, at first, it is necessary that the cascade file is trained separately by OpenCV (open source computer vision) software and we should provide a XML file of it. OpenCV software is used in classification to detect the object. The vehicle images in dataset are given to this software to make training and XML files. Then, we can detect vehicle by using Haar-Like approach.

For vehicle detection, dataset of different images of vehicle is the first thing that we need. We can use the datasets which are available on the internet, however, the dataset used in this paper has been provided by surveillance videos. The database consists of 576 images in different sizes and different vehicles in different light conditions. See a few samples of the database that we are use in vehicle detection as Fig. 4.

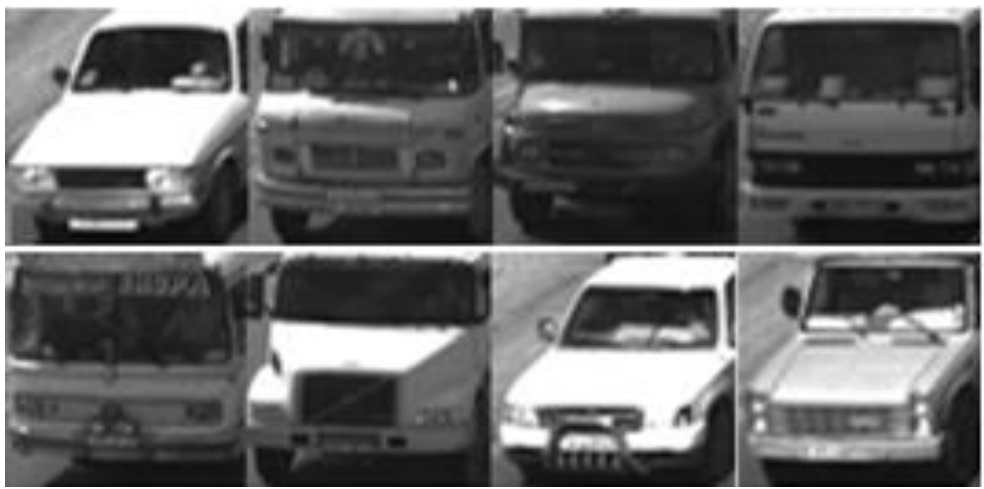

**Figure 4.** Some samples of the used dataset.

## 3. PREPARING XML FILES FOR CAR IDENTIFICATION

For our classifier training, we need a set of samples. There are two types of samples: positive and negative. Positive samples are corresponded to the images with frontal view of vehicles. Negative samples are corresponded to non-vehicle images. The next step is the training of classifier. OpenComputer Vision offers two different applications for training a Haar classifier: "opencv_haartraining" and "opencv_traincascade", but only the newer "opencv_traincascade" will be described further. It allows the training process to be multi-threaded for reducing the processing time. The training cascade generates an XML file when the process is completely finished. So after this, we can identify vehicle by viola Jones algorithm.

By considering this point that the database in this paper consists of vehicles from the frontal view, cars are detected well. If we want to detect the vehicles which are in the back side or aside, we should provide another database for that side and repeat the steps.

## 4. EXPERIMENTAL RESULT

In this paper, all used images are from real scene traffic in different light positions, the videos' formats are AVI. In order to prove the effectiveness of algorithm, we use Matlab and OpenCV to simulate the experiment.

### 4.1. Evaluation Metrics

In order to evaluate the vehicle detection results, numerical accuracy assessment has been done by comparing the number of vehicles identified manually and automatically by the proposed method. Three types of numerical metrics for the extracted results are defined as Eq. (1) to Eq. (3).

$$Accuracy = \frac{TP}{TP + FP} \quad (1)$$

$$Completeness = \frac{TP}{TP + FN} \quad (2)$$

$$Quality = \frac{TP}{TP + FP + FN} \quad (3)$$

Where; True Positive (TP) is the number of true vehicles which are correctly extracted, False Positive (FP) is the number of false vehicles and False Negative (FN) is the number of omitted vehicles (missed). *Accuracy* metric means detection accuracy rate in comparison to the ground truth, *Completeness* means how many of vehicles are correctly detected and the *Quality* shows the overall accuracy of extraction methods.

### 4.2. Evaluation Measures

The obtained results of our approach in various scenes of surveillance videos under strong, weak and medium light conditions are shown in Table 1.

table I. EXPERIMENTAL RESULTS.

| Scenes | Accuracy | Completeness | Quality |
|---|---|---|---|
| Scene 1 | 94% | 96% | 90% |
| Scene 2 | 95% | 97% | 92% |
| Scene 3 | 93% | 94% | 88% |
| Scene 4 | 92% | 82% | 77% |

Many video scenes under different light condition are used to detect vehicles and show the algorithm performance. The sequences used in our experiments are summarized in Table 2. Scenes 1, 2 and 3 are real surveillance videos from Isfahan city in different lighting conditions and "Scene 4" is a standard Highway video from the ATON video benchmark set as the test video which is named Haighway1 in here. Fig. 5 shows the vehicle detection results in medium light condition (the name of this video is "Scene 1"). Image (a) in Fig. 5 shows sedan and truck detection results and images (b) to (f) show the detection result of different vehicles in different distances of camera, the proposed algorithm detects all of them.

table II. IMAGE SEQUENCES USED IN OUR EVALUATIONS.

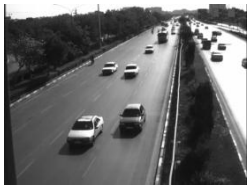 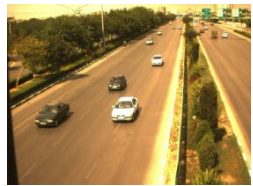 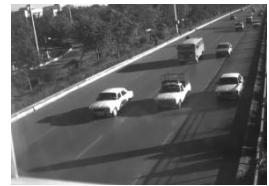 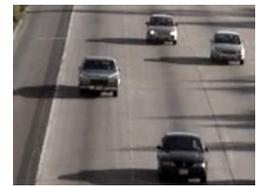

| Labeled scenes | Scene 1 | Scene 2 | Scene 3 | Scene 4 |
|---|---|---|---|---|
| Frames Number | 3479 | 2171 | 1131 | 440 |
| Frame Size | 640×480 | 640×480 | 640×480 | 320×240 |
| Frame rate | 15 | 30 | 15 | 15 |
| Scene Type | Outdoor | Outdoor | Outdoor | Outdoor |
| Surface | Asphalt | Clear Asphalt | Asphalt | Asphalt |
| Noise | Medium | Low | Medium | High |
| Lighting | Medium | High | Low | Medium |
| Shadow Size | Low | Very low | High | Medium |
| Shadow Strength | High | High | High | High |
| Shadow Direction | Horizontal/Front | Under vehicles | Horizontal/Left | Horizontal/Left |

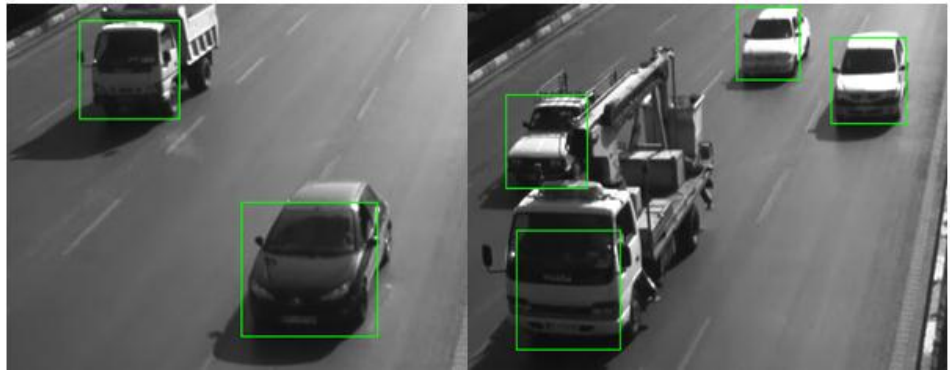

(a) (b)

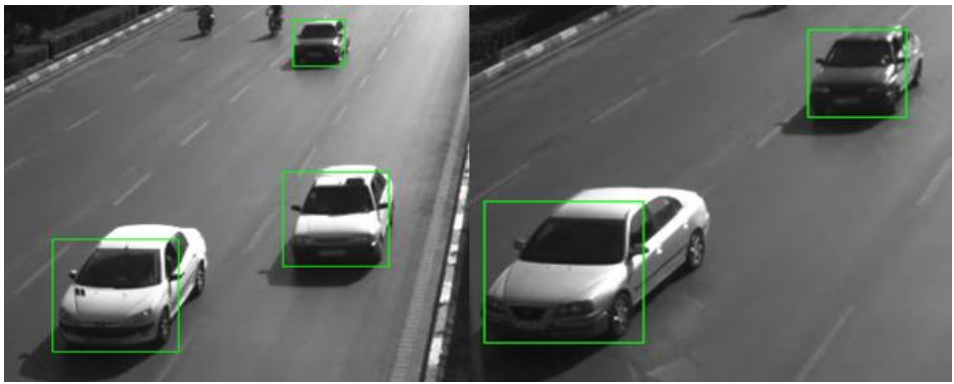

(c) (d)

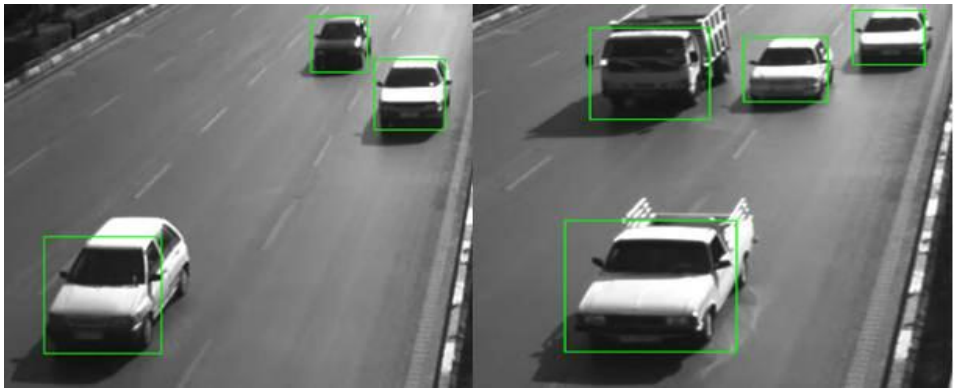

(e) (f)

**Figure 5.** Vehicle detection results in medium lighting.

Fig. 6 shows the detection result in strong light condition, this video is named "Scene 2". Images (a) and (b) show the detection result of different vehicles at the same time, image (c) shows detection result of a truck with two sedans, image (d) shows the result detection of bus and image (e) shows three sedans detection concurrently. It is observable that the used algorithm is able to detect vehicles separately in this light condition.

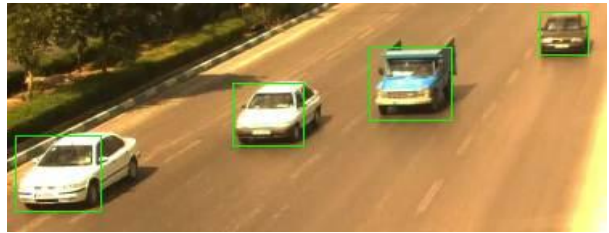

(a)

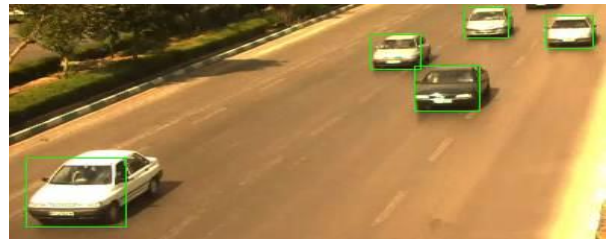

(b)

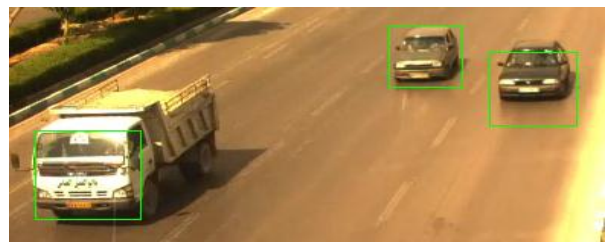

(c)

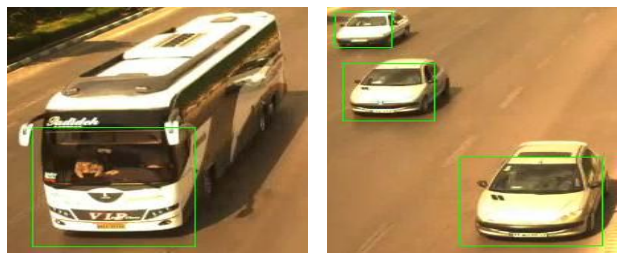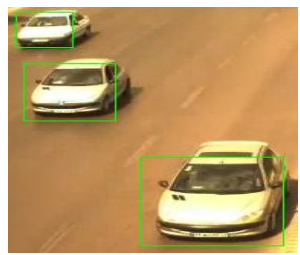

(d) (e)

**Figure 6.** Vehicle detection results in strong lighting.

Fig. 7 shows the vehicle detection results in weak lighting condition; this video is named "Scene 3". Image (a) shows the detection result of a van and sedan at the same time, images (b) and (c) shows the result of the detection of a few vehicles in different colors and sizes, image (d) shows the detection result of a truck in weak light condition.

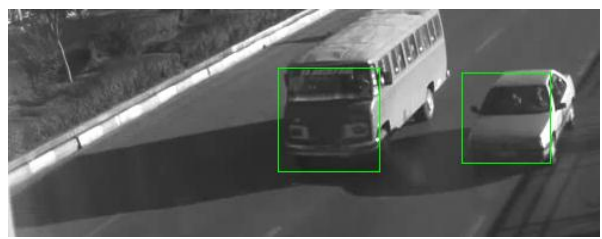

(a)

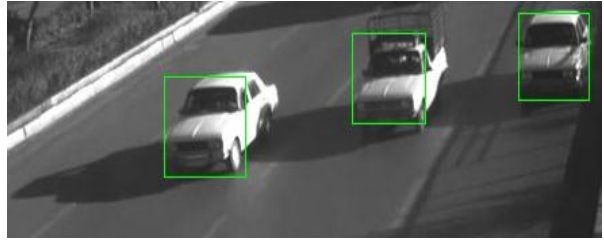

(b)

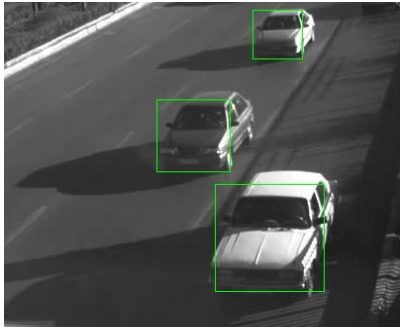
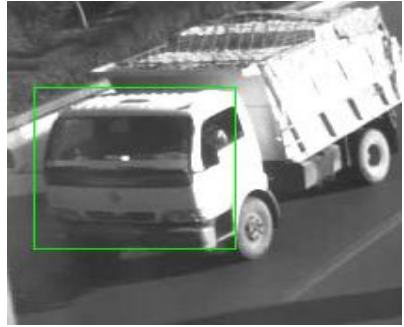

(c)　　　　　　　　　　　　　　(d)

**Figure 7.** Vehicle detection results in low lighting.

### 4.3. Comparison to Other Algorithms

To test the effect of car detection based on the proposed algorithm along with other algorithms, we use the value of accuracy for all algorithms. The result is shown in Fig. 8.

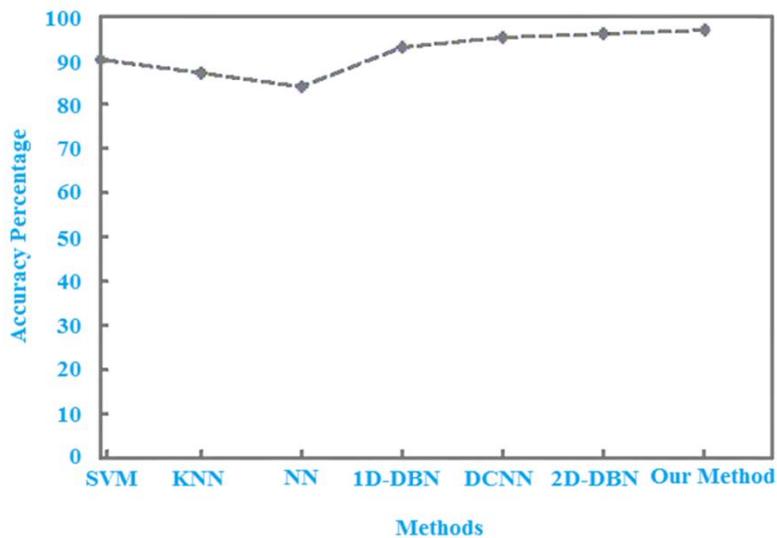

**Figure 8.** Accuracy (%) results in different method type.

Fig. 8 shows the Accuracy of vehicle detection in different methods. From the compared results, the method proposed is better than the others. According to the obtained vehicle detection results in Table 1, the accuracy, completeness, and quality rates of the proposed method were about 94%, 92%, and 87%, respectively, in all lighting conditions.

The obtain results by the proposed algorithm shows the efficiency of this algorithm in detecting different vehicles in different light conditions.

## 5. CONCLOTIONS

In this paper, we proposed a real-time method for detecting different vehicles by Viola Jones algorithm which is based on AdaBoost classification technique. AdaBoost combines some weak classifiers to make a strong classifier. The obtained results in real surveillance videos from Isfahan city in different lighting conditions show the efficiency and the practicality of this method. We can use the obtain results in other computer vision application such as registering the driving violation and vehicle plate registration or other related aspects of complex systems.